\documentclass{article}

\PassOptionsToPackage{numbers, compress}{natbib}

\usepackage[final]{neurips_2021_ml4ps}

\usepackage[utf8]{inputenc} 
\usepackage[T1]{fontenc}    
\usepackage{hyperref}       
\usepackage{url}            
\usepackage{booktabs}       
\usepackage{amsfonts}       
\usepackage{nicefrac}       
\usepackage{microtype}      
\usepackage{xcolor}         

\usepackage{amsmath}
\usepackage{graphicx}

\usepackage{tikz}
\usetikzlibrary{matrix,
                positioning,
                quotes,
                cd}

\title{A posteriori learning of quasi-geostrophic turbulence parametrization: an experiment on integration steps}

\author{%
Hugo Frezat\\
Univ. Grenoble Alpes, CNRS UMR LEGI, Grenoble, France\\
Univ. Grenoble Alpes, CNRS UMR IGE, Grenoble, France\\
IMT Atlantique, CNRS UMR Lab-STICC, Brest, France\\
\texttt{hugo.frezat@univ-grenoble-alpes.fr} \And
Julien Le Sommer\\
Univ. Grenoble Alpes, CNRS UMR IGE, Grenoble, France \And
Ronan Fablet\\
IMT Atlantique, CNRS UMR Lab-STICC, Brest, France \And
Guillaume Balarac\\
Univ. Grenoble Alpes, CNRS UMR LEGI, Grenoble, France\\
Institut Universitaire de France (IUF), Paris, France \And
Redouane Lguensat\\
Learning, Data and Robotics Lab, ESIEA, Paris, France\\
LOCEAN-IPSL, Sorbonne Université, Institut Pierre Simon Laplace, Paris, France
}

\begin{document}

\maketitle

\begin{abstract}
  Modeling the subgrid-scale dynamics of reduced models is a long standing open problem that finds application in ocean, atmosphere and climate predictions where direct numerical simulation (DNS) is impossible. While neural networks (NNs) have already been applied to a range of three-dimensional flows with success, two dimensional flows are more challenging because of the backscatter of energy from small to large scales. We show that learning a model jointly with the dynamical solver and a meaningful \textit{a posteriori}-based loss function lead to stable and realistic simulations when applied to quasi-geostrophic turbulence.
\end{abstract}

\section{Introduction}
Understanding and predicting the evolution of various natural systems would not be possible without simulations of turbulent flows. However, solving all the spatial and temporal scales of the associated partial differential equation (PDE), i.e., the Navier-Stokes equations remains computationally prohibitive. One popular solution (e.g. \cite{meneveau2000scale, graham2013framework, dipankar2015large}) is to resolve only the largest scales and use subgrid closures (or physical parametrizations) to represent the smaller ones. 

Recently, neural networks (NNs) have been proposed as a promising alternative to algebraic parametrizations in three-dimensional incompressible turbulence \cite{gamahara2017searching, park2021toward, frezat2021physical}. Two-dimensional problems, however, are more challenging due to the inverse cascade of energy that leads to negative viscosities, and it has been demonstrated that numerical stability of the trained model in decaying turbulence requires either the removal of negative eddy viscosities \cite{maulik2019subgrid} or a large training dataset \cite{guan2021stable}.
2D turbulence is also a simple model that well approximates more complex rotating stratified flows (found in atmosphere and ocean dynamics).

\begin{figure}
    \centering
    \tikzcdset{nodes={font=\scriptsize}} 
    \begin{tikzcd}
    \mathbf{y}(t) \arrow[rr, "f(\mathbf{y}(t))" description] \arrow[dd, "\mathcal{T}(\mathbf{y}(t))" description] & & \mathbf{y}(t + \Delta t) \arrow[r, phantom, "\cdots"] & \mathbf{y}(t + N \Delta t) \arrow[rr, "f(\mathbf{y}(t + N \Delta t))" description] \arrow[ddl, blue, phantom, shift right=.3ex, "\cdots"] & & \mathbf{y}(t + (N + 1) \Delta t) \\
    & g(\bar{\mathbf{y}}(t)) \arrow[d] & & & g(\bar{\mathbf{y}}(t + N \Delta t)) \arrow[d] \\
    \bar{\mathbf{y}}(t) \arrow[ru, bend left=20] \arrow[rd, bend right=20] & + \arrow[r] & \bar{\mathbf{y}}(t + \Delta t) \arrow[uu, <->, dashed, blue, "\mathcal{L}_{\mathrm{a posteriori}}" description] \arrow[r, phantom, "\cdots"] & \bar{\mathbf{y}}(t + N \Delta t) \arrow[ru, bend left=20] \arrow[rd, bend right=20] \arrow[uu, <->, dashed, blue, "\mathcal{L}_{\mathrm{a posteriori}}" description] & + \arrow[r] & \bar{\mathbf{y}}(t + (N + 1) \Delta t) \arrow[uu, <->, dashed, blue, "\mathcal{L}_{\mathrm{a posteriori}}" description] \\
    & \mathcal{M}_{\mathrm{NN}}(\bar{\mathbf{y}}(t)) \arrow[u] \arrow[d, <->, dashed, red, "\mathcal{L}_{\mathrm{a priori}}"] & & & \mathcal{M}_{\mathrm{NN}}(\bar{\mathbf{y}}(t + N \Delta t)) \arrow[u] \\
    & R(\bar{\mathbf{y}}(t))
    \end{tikzcd}
    \caption{Sketch of one learning step for the presented strategies. The \textit{a priori} loss is computed at instantaneous time $t$ (dashed, red), while the \textit{a posteriori} loss is evaluated on states forward in time (dashed, blue).}
    \label{fig:sketch_learning}
\end{figure}
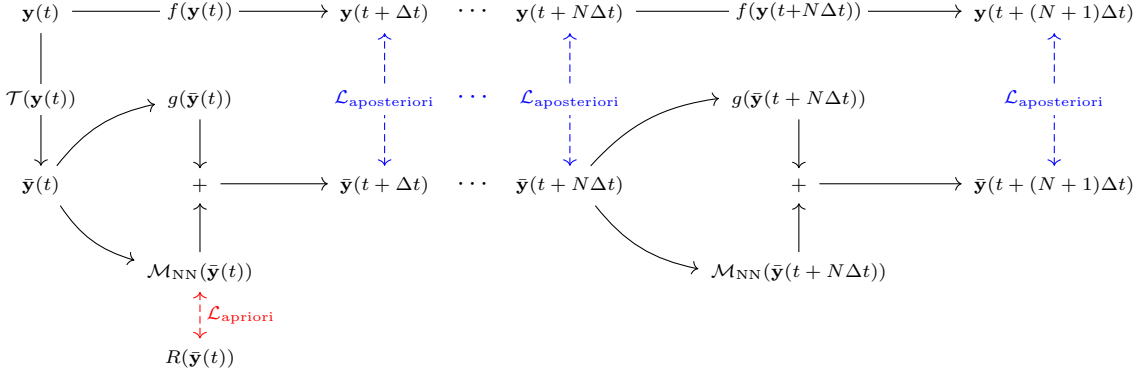

We show that on a forced configuration of 2D turbulence, stability can be obtained when the NN is trained end-to-end on a temporal horizon, i.e. jointly with the forward solver (see Fig. \ref{fig:sketch_learning}). At equivalent computational complexity, the standard learning strategy does not lead to a stable parametrization and is thus not usable in practice. Additionally, the presented strategy allows us to constrain the calibration of the model on \textit{a posteriori} metrics, i.e. the evolution of quantities of interest over time, which are the absolute performance measures in turbulence modeling. Finally, we show that the number of iterations performed during the training process have a significant impact on the long-time statistics of the model.

\section{A priori and a posteriori learning}
In the machine learning community, it is well known that models can take advantage of additional informations during training. For instance, the basic, purely data-based approach has been extended to train models end-to-end using an entire system. In practice, some examples of those systems include autonomous vehicles \cite{bojarski2016end}, objects detection \cite{zhou2018voxelnet} and super-resolution \cite{chen2018fsrnet}. This idea has also recently found many applications in the physical sciences community (e.g. \cite{holl2019learning, kochkov2021machine}).

In the context of partial differential equation parametrization, we are given with a high-resolution system $f(\mathbf{y})$ and a low resolution system $g(\mathbf{\bar{y}})$ describing the evolution of high and low resolution variables $\mathbf{y}(t)$ and $\bar{\mathbf{y}}(t)$ respectively by,
\begin{align}
    \begin{cases}
    \dfrac{\partial \mathbf{y}}{\partial t}        = f(\mathbf{y}), \hspace{5mm} &\mathbf{y} \in \Omega \\[2ex]
    \dfrac{\partial \bar{\mathbf{y}}}{\partial t}  = g(\bar{\mathbf{y}}) + R(\bar{\mathbf{y}}), \hspace{5mm} &\bar{\mathbf{y}} \in \bar{\Omega} \\[2ex]
    \mathcal{T}(\mathbf{y}) = \bar{\mathbf{y}}\\
    \end{cases}
\end{align}
where $\bar{\Omega}$ is a smaller domain than $\Omega$ and $\mathcal{T}$ is a known projection operator that goes from high to low resolution variables.

In this framework, we are interested in modeling the residual term, or parametrization $R(\bar{\mathbf{y}})$ with a neural network $\mathcal{M}_{\mathrm{NN}}(\bar{\mathbf{y}})$, based on high resolution data coming from a direct numerical simulation of $f(\mathbf{y})$, collected in a dataset
\begin{equation}
    \mathcal{D} := \{\bar{\mathbf{y}}(t)\} \rightarrow \{R(\bar{\mathbf{y}}(t))\}.
\end{equation}
If we connect the machine learning naming to fluid dynamics concepts, we could say that the data-based approach only gives the ability to optimize a model on \textit{a priori} metrics, i.e., instantaneous predictions of the targeted term. With the end-to-end approach, however, we have access to simulation quantities over time, which can be used to train the model on \textit{a posteriori} metrics (See \cite{pope2001turbulent} for more details). 
Using the data $\mathcal{D}$, a standard learning strategy based on the direct prediction of the residual term can be written as $\operatorname*{arg\,min}_\theta \mathcal{L}(R(\bar{\mathbf{y}}), \mathcal{M}_{\mathrm{NN}}(\bar{\mathbf{y}} | \theta))$, which has already been applied to fluid dynamics parametrization, and corresponds to an \textit{a priori} minimization problem.
Now, if we have a differentiable solver for $g(\bar{\textbf{y}})$, the standard minimization problem performed during learning can be rewritten as a series of \textit{a posteriori} losses on low resolution variables that are advanced in time, such that
\begin{equation}
    \operatorname*{arg\,min}_\theta \mathcal{L}(\mathbf{y}(t), \bar{\mathbf{y}}(t_{0}) + \Phi_{\theta}(\bar{\mathbf{y}}(t_{0}), t_{0}, t)), t \in [0, T]
    \label{eq:min_e2e}
\end{equation}
where $\mathcal{L}$ is the loss function, $[0, T]$ is the temporal horizon on which the forward solver is integrated and $\Phi$ the flow operator that advances the reduced model in time, expressed as
\begin{equation}
    \Phi_{\theta}(\bar{\mathbf{y}}, t_{0}, t) = \int_{t_{0}}^{t} g(\bar{\mathbf{y}}(\tau)) + \mathcal{M}_{\mathrm{NN}}(\bar{\mathbf{y}}(\tau) | \theta) \, \mathrm{d} \tau.
\end{equation}
In practice, the forward solver performs $N$ discrete timesteps, which defines the temporal horizon $T = N \Delta t$. A sketch of the \textit{a posteriori} learning strategy is shown in Fig. \ref{fig:sketch_learning}.

\section{Application to quasi-geostrophic turbulence parametrization}
The described strategies are applied to a classical two-dimensional single-layer barotropic quasi-geostrophic (QG) potential vorticity equation \cite{majda2006nonlinear}, defined by
\begin{equation}
    \partial_{t}\omega + J(\psi, \omega) = \nu \nabla^{2} \omega - \mu \omega + F
    \label{eq:qg}
\end{equation}
with
\begin{align}
    \mathbf{u} &= (-\partial_{y}\psi, \partial_{x}\psi)\\
    \omega &= \nabla^{2} \psi
    \label{eq:flow}
\end{align}
where $\omega$ is the vorticity, $\psi$ the stream function, $\mathbf{u}$ the velocity vector and $J(\psi, \omega) = \partial_{x}\psi \partial_{y}\omega - \partial_{y}\psi \partial_{x}\omega$ is the Jacobian operator. The model is parametrized by a viscosity $\nu$, a linear drag $\mu$ and a source term $F$.

\textbf{Projecting to the reduced system}. In turbulence modeling, a common approach is to suppose that the reduced system has been filtered so that the smallest scales are removed, and instead predicted by a model. In practice, the projection operator $\mathcal{T}$ is a spatial filter $\Lambda_{\delta}(x, y)$ at spatial scale $\delta > 0$ defined such that
\begin{equation}
    \mathcal{T}(\omega) := \omega * \Lambda_{\delta} = \bar{\omega}(x, y, t)
\end{equation}
and the reduced system becomes 
\begin{equation}
    \partial_{t}\bar{\omega} + J(\bar{\psi}, \bar{\omega}) = \nu \nabla^{2} \bar{\omega} - \mu \bar{\omega} + \bar{F} + \underbrace{J(\bar{\psi}, \bar{\omega}) - \overline{J(\psi, \omega)}}_{R(\psi, \omega)}
\end{equation}
where $R(\psi, \omega)$ is the subgrid-scale (SGS) term or parametrization that needs to be modeled.

\textbf{Configuration.} In our experiment, we discretize the full solutions on $2048 \times 2048$ grid points using a pseudo-spectral method and use $\delta = 16$ with a spectral cut-off filter $\Lambda_{\delta}(k) = 0, \forall k > \pi / \delta \Delta$ where $\Delta$ is the filter width so that our reduced solution lies on a $128 \times 128$ grid. The source term $F$ is a time-dependent wind-forcing acting at large scales $k = 4$, 
\begin{equation}
    F = C_{F}(t) \left[ \cos(4y + \pi \sin(1.4t)) - \cos(4x + \pi \sin(1.5t)) \right]
\end{equation}
with steady enstrophy rate $C_{F}$ such that $\langle F^2 \rangle / 2 = 3$.
Following the procedure given by \cite{graham2013framework}, we spin-up multiple simulations with $1024 \times 1024$ grid initialized with random modes at $k = 4$ for approximately $1300$ days. 
The simulations are then integrated for $35$ days on the high-resolution grid to obtain a turbulent state that will be used as initial conditions for training and testing. Finally, training data is extracted from 10 independent $\approx 65$ days long ($48000$ iterations) trajectories. The parameters of the simulations are summarized in dimensional and non-dimensional units in Table \ref{tbl:parameters}.

\begin{table}
    \label{tbl:parameters}
    \caption{Parameters of the forced configuration in dimensional and non-dimensional units, with $L_{d}(x) = 1.2 \times 10^6 s$ and $T_{d}(x) = 504 \times 10^4 / \pi m$.}
    \centering
    \begin{tabular}{l c c}
        \toprule
         Parameter & Dimensional & Numerical \\
        \hline
        Domain length ($L$) & $10^4$ km & $2\pi$ \\
        Spatial resolution ($\Delta x$) & $10$ km & $3.00 \times 10^{-3}$ \\
        Time step ($\Delta t$) & $120$ s & $1.00 \times 10^{-4}$\\
        Linear drag ($\mu$) & $1.25 \times 10^{-8}$ m$^{-1}$ & $2.00 \times 10^{-2}$\\
        Viscosity ($\nu$) & $22.0$ m$^2$ s$^{-1}$ & $1.02 \times 10^{-5}$\\
        Reynolds number ($Re$) & \multicolumn{2}{c}{$\approx 220000$}\\
        \bottomrule
    \end{tabular}
\end{table}

Our model \footnote{\url{https://github.com/hrkz/torchqg}} is implemented with PyTorch \cite{paszke2019pytorch} in order to benefit from automatic differentiation, which is required when computing the gradient of the flow operator \eqref{eq:flow}.

\textbf{Learning components.} The neural network $\mathcal{M}_{\mathrm{NN}}$ used in this experiment for both \textit{a priori} and \textit{a posteriori} strategies is realized with a fully convolutional architecture.  It consist of 10 convolutional layers with 64 features each, interspresed with ReLU activation functions using kernel sizes of $5 \times 5$. The parameters are optimized for a fixed number of epochs with Adam and a learning rate of $10^{-4}$.

In particular, the \textit{a posteriori} version is trained for a different number of iterations $N = \{1, 5, 30\}$, with a loss function that minimizes the mean squared error (MSE) of vorticity $\bar{\omega}$ over time, such that
\begin{equation}
    \mathcal{L}_{\mathrm{a posteriori}} := \frac{1}{N} \sum_{i=1}^{N}(\omega(i \Delta t) * \Lambda_{\delta} - \bar{\omega}(i \Delta t))^2
\end{equation}

\textbf{Results.} We run a new simulation at $\delta = 16$, i.e. on a $128 \times 128$ grid, comparing the NNs to the reference direct numerical simulation $\overline{\mathrm{DNS}}$ and the dynamic Smagorinsky model \cite{smagorinsky1963general,germano1991dynamic}, a purely dissipative algebraic model which is a common baseline in the fluid dynamics community. The run is performed for 3000 and 48000 iterations of the low resolution and direct numerical simulation systems, respectively, since the timestep for low resolution systems can be $\delta$ times larger than the DNS $\Delta t$.

\begin{figure}
\centering
\includegraphics[width=\textwidth]{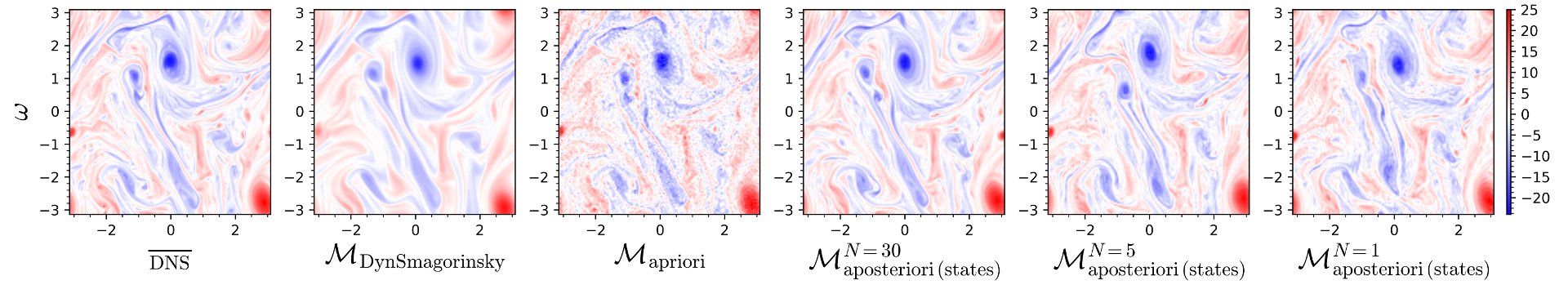}
\caption{Vorticity fields at the end of the testing simulation with the filtered DNS (leftmost) and the discussed models. Note that the model trained with \textit{a priori} learning was numerically unstable and its states diverged to infinity.}
\label{fig:fields}
\end{figure}

First, we can see in Fig. \ref{fig:fields} the vorticity fields $\omega$ at the end of the new simulation. As expected, the dynamic Smagorinsky model is only dissipative and thus the smallest scales of the simulation are not visible anymore. On the other side, the model learnt from the \textit{a priori} strategy accumulated too much energy on the smallest scales which led to numerical instabilities, as expected from similar studies \cite{maulik2019subgrid, guan2021stable}. The 3 models trained with the \textit{a posteriori} strategy remain stable at the end of the simulation. Moreover, these models are preserving the small-scale features of the flow. Now, we look at quadratic invariants of 2D turbulence, which develops as a double cascade in statistically stationary conditions \cite{boffetta2007energy}. In particular, the enstrophy wavenumber spectrum $Z(k)$ and flux $\Pi_{Z}(k)$ are relevant measures of the success of turbulence parametrizations, as proposed by \cite{graham2013framework}.

\begin{figure}
\centering
\includegraphics[width=0.8\textwidth]{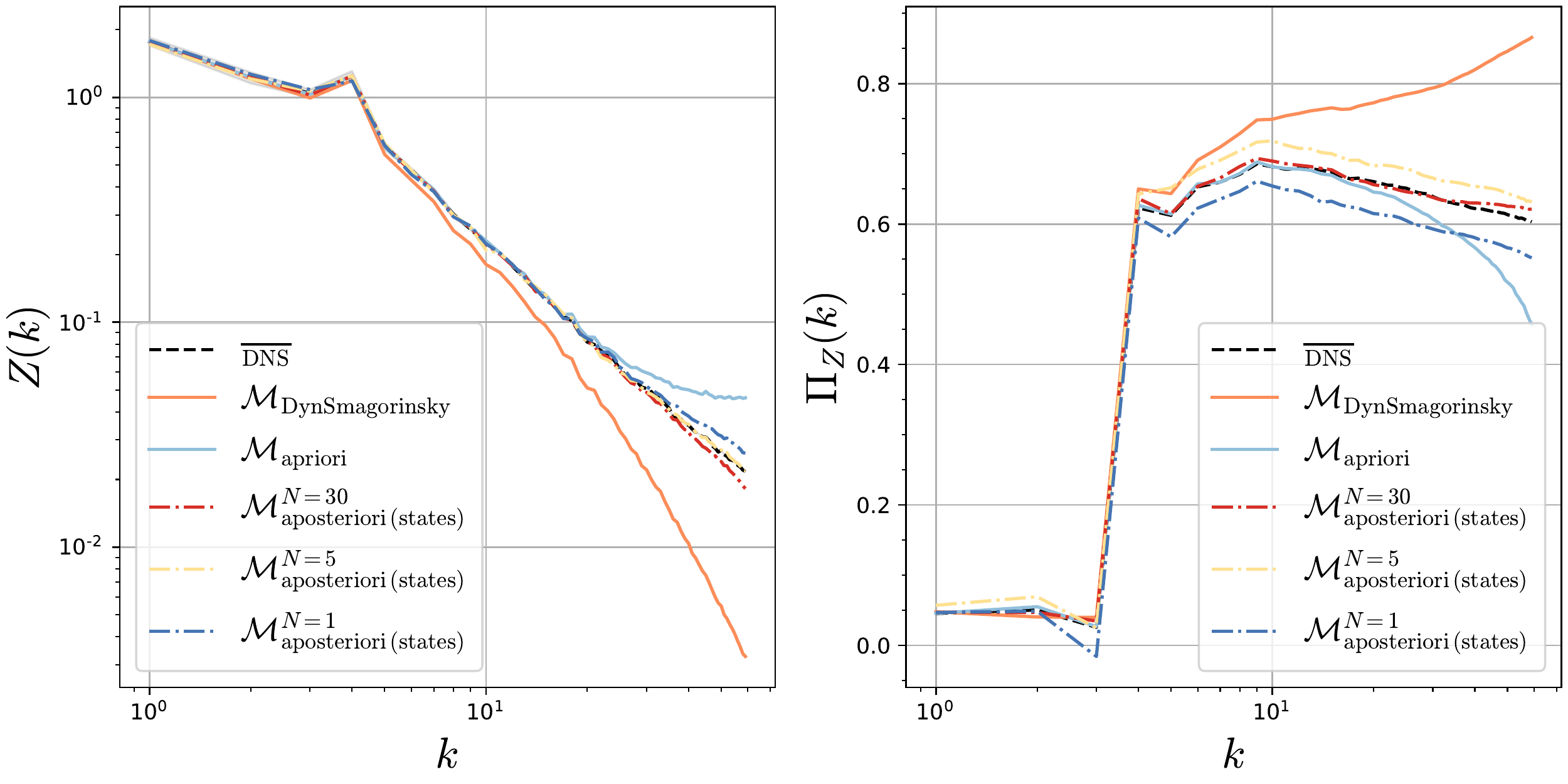}
\caption{Enstrophy spectrum (left) and time-averaged enstrophy flux (right) of the different models w.r.t the reference DNS (dashed). For the NNs models, only those trained with the \textit{a posteriori} strategy are stable and thus have valid statistics.}
\label{fig:stats}
\end{figure}

In Fig. \ref{fig:stats} (left), the enstrophy spectra confirm a large deviation at the smallest scales (highest wavenumbers $k$) for the Smagorinsky model, and a visible deviation at the largest scales (smallest wavenumbers $k$) for the \textit{a priori} model and \textit{a posteriori} model trained with $N = 1$. Overall statistical performance of the model is verified by the enstrophy flux in Fig. \ref{fig:stats} (right), where the NN trained with the \textit{a posteriori} strategy at $N = 30$ is the closest to the exact flux predicted by the DNS.

\section{Conclusion}
We show that a NN trained jointly with the forward solver on \textit{a posteriori} informations lead to stable simulations and outperforms the classical data-driven strategy and algebraic models based on different statistical metrics for the prediction of quasi-geostrophic turbulence parametrization. However, it is important to consider that this requires a differentiable solver for the reduced order system, which is not the case for most global circulation models developed nowadays.

\subsection*{Acknowledgments}
The authors thank Laure Zanna, Olivier Pannekoucke and Corentin Lapeyre for the helpful discussions. This research was supported by the CNRS through the 80 PRIME project and the ANR through the Melody, OceaniX, and HRMES projects. Additional support was provided by Schmidt Futures, a philanthropic initiative founded by Eric and Wendy Schmidt, as part of its Virtual Earth System Research Institute (VESRI). Computations were performed using GPU resources from GENCI-IDRIS.

\bibliographystyle{plainnat}
{\footnotesize\bibliography{main}}

\end{document}